\documentclass[10pt]{article}
\newif\ifblindreview

\usepackage[letterpaper]{geometry}
\usepackage{amta2024}
\usepackage{times}
\usepackage{url}
\usepackage{latexsym}
\usepackage{natbib}
\usepackage{layout}
\usepackage{multicol}
\usepackage{booktabs,array}
\usepackage{float}
\usepackage[hidelinks]{hyperref}
\usepackage[all]{hypcap}
\usepackage{graphicx}
\usepackage{inconsolata}
\usepackage{xcolor}
\hypersetup{
    colorlinks,
    linkcolor={red!50!black},
    citecolor={blue!50!black},
    urlcolor={blue!80!black}
}

\ifblindreview
  \newcommand{\authorinfo}{\author{}} 
\else
  \newcommand{\authorinfo}{
    \author{
            \name{\bf Richard Yue} \hfill \addr{yue.r@northeastern.edu}\\
            \addr{\small Northeastern University, San Jose, CA}
    \AND
           \name{\bf John E. Ortega} \hfill \addr{j.ortega@northeastern.edu}\\
            \addr{\small Institute for Experiential AI, Northeastern University, Boston, MA}
    }
  }
\fi

\usepackage{todonotes}

\begin{document}

\amtaHeader{x}{x}{xxx-xxx}{2015}{45-character paper description goes here}{Author(s) initials and last name go here}
\title{Predicting Anchored Text from Translation Memories for Machine Translation Using Deep Learning Methods}
\authorinfo

\maketitle
\pagestyle{empty}

\begin{abstract}
\vspace{5pt}
  Translation memories (TMs) are the backbone for professional translation tools called computer-aided translation (CAT) tools. In order to perform a translation using a CAT tool, a translator uses the TM to gather translations similar to the desired segment to translate ($s'$). Many CAT tools offer a fuzzy-match algorithm to locate segments ($s$) in the TM that are close in distance to $s'$. After locating two similar segments, the CAT tool will present parallel segments ($s$, $t$) that contain one segment in the source language along with its translation in the target language. Additionally, CAT tools contain fuzzy-match repair (FMR) techniques that will automatically use the parallel segments from the TM to create new TM entries containing a modified version of the original with the idea in mind that it will be the translation of $s'$. Most FMR techniques use machine translation as a way of ``repairing'' those words that have to be modified. In this article, we show that for a large part of those words which are \textit{anchored}, we can use other techniques that are based on machine learning approaches such as Word2Vec. BERT, and even ChatGPT. Specifically, we show that for anchored words that follow the continuous bag-of-words (CBOW) paradigm, Word2Vec, BERT, and GPT-4 can be used to achieve similar and, for some cases, better results than neural machine translation for translating anchored words from French to English.
\end{abstract}

\begin{multicols}{2}

\section{Introduction}\label{se:intro}
Professional translators use computer-aided translation (CAT) tools \citep{bowker2002computer} to translate text from one language called the source language (SL) to a target language (TL). Most CAT tools have an option known as \textit{fuzzy-match repair} (FMR)  \citep{kranias04:_automatic,hewavitharana2005augmenting,dandapat2011,ortega16a,bulte18:_m3tra,tezcan2021towards}, which is backed by a parallel translation memory (TM) that contains sentences (called segments) in the SL and TL. Each pair, or unit, of parallel segments in the TM is known as a \textit{translation unit} (TU). A TU contains a source segment ($s$) along with a target segment ($t$). When a professional translator attempts to translate a segment in the SL (denoted as $s'$) a fuzzy-match lookup is performed using a word-based Levenshtein distance \citep{levenshtein1965binary} between $s'$ and $s$ where a 100\% match means that the words from $s'$ are identical to the words in $s$. It is often the case that a professional translator uses matches from FMR to only translate a few words (called sub-segments) from the entire segment. In this article, we focus on improving those cases where there exists only one word to translate, known as an \textit{anchored} word, whose position is in between two words that are already captured. In our studies, the anchored word is a common case that professional translators often use. We experiment with four techniques to translate the anchored word: (1) Neural Machine Translation, (2) a BERT-based \citep{sanh2019distilbert} implementation, (3) Word2Vec \citep{mikolov2013efficient} and (4) OpenAI GPT-4 prompting \citep{achiam2023gpt}.

The prediction of an anchored word has been presented in many contexts and can be considered the main objective of a language model. Several models based on attention allow a weight to be assigned to certain words within a context window so that surrounding words that strongly influence the overall context can have a greater impact on the prediction made. This could potentially be used in order to improve predictions made for anchored text by taking longer contexts into account than the surrounding words. We discuss this approach in the context of generative models, where such systems could be harnessed to generate highly accurate predictions.

The rest of this article is structured as follows.  The next section discusses related work by accentuating the differences between FMR based on MT and anchored-word prediction. Section \ref{se:methodology} then presents the BERT, Word2Vec, and GPT-4 approaches used for translating anchored words. In Section \ref{se:experimental}, we describe the corpus and configurations used for our experiments whose results are reflected in Section \ref{se:results}, followed by concluding remarks in Section \ref{se:conclusion}.

\section{Related Work}\label{se:related}

For the majority of FMR approaches, MT is used to translate mismatches, regardless if they are anchored words or not. Generally, MT techniques for FMR are focused on the decoding process where statistical-based systems \citep{biccici2008dynamic,simard2009phrase,zhechev2010seeding,koehn2010convergence,li16j,liu19} or neural-based systems \citep{ortega14a,ortega16a,gu18,bulte18:_m3tra,bulte-tezcan-2019-neural} are used in such a manner to ``repair'' either the MT system or the mismatched sub-segments between $s'$ and $s$. This article is focused on repairing the mismatched sub-segments in specific situations where sub-segments of $s$ are common in $s'$ with the exception of one word (e.g. $s$=`the \textbf{brown} dog' and $s'$=`the \textbf{red} dog').

Previous work \citep{ortega16a,bulte18:_m3tra} can be considered identical to this article as it uses FMR to first find mismatches between $s'$ and $s$ and then translates the missing words with different MT systems. However, their system uses context around \textit{all} mismatches where we only consider mismatches with anchored words, similar to \cite{kranias04:_automatic}. While other techniques \citep{hewavitharana2005augmenting,dandapat2011} are based on probabilistic MT models or employ different algorithms for aligning $s'$ and $s$, we use a word-based edit distance \citep{levenshtein1965binary,Wagner:1974:SCP:321796.321811} that marks the mismatched sub-segments and discards non-anchored words.

\cite{Tezcan2022EvaluatingTI} investigate a wide range of automatic quality estimation (QE) metrics in order to assess what effect integrating fuzzy matches can have on a number of aspects of translation quality, in addition to performing manual MT error analysis. They further evaluate what influence fuzzy matches have on a translation and how further quality improvements can be made by quantitative analyses that focus on the specific characteristics of a retrieved fuzzy match. Neural Fuzzy Repair (NFR) outperforms baselines in all automated evaluation metrics. There was not a discernable difference between NFR and Neural Machine Translation (NMT) error in manual evaluation, but different error profiles emerged in this study, highlighting some of the strengths and weaknesses of each method. Namely, NFR produced more errors in the category of ``semantically unrelated", whereas the baseline NMT system produced more errors in the categories of ``word sense" and ``multi-word expression". The NFR system made more accuracy errors, but producing fluent output was its strong suit. Meanwhile, in terms of lexical choices, NMT produced more ``non-existing/foreign" errors, which was not an issue for NFR. The baseline system performed better on grammar and syntax. Our study differs in that it focuses specifically on anchored text and on leveraging the strengths of language models in next word prediction in order to fill in single-word gaps.

\cite{EsplGomis2011UsingWA} attempt to improve CAT via the TM using pre-computed word alignments between source and target TUs in the TM. When a user is translating $s'$ with a fuzzy match score greater than or equal to 60\%, the proposed system marks the words that need to change as well as those that must remain the same in order to obtain $t'$. Alignments are obtained from GIZA++ \citep{brown93:tmo,vogel96:hbw} and take both a statistical and syntactic approach to detecting where changes need to occur. The experiments offer insight into how human decisions to keep/change text during translation can be integrated into FMR. Our approach differs in that we specifically locate anchored text and, following that, continue on to a prediction step, providing the content needed to perform fuzzy match repair in the translation step.

\cite{irsoy2020corrected} compare performance of pre-trained word embeddings in use in language models such as BERT with continuous bag of words (CBOW) embeddings trained with Word2Vec. The authors claim that, while BERT embeddings are useful and effective, they often offer only marginal gains as compared to Word2Vec embeddings trained using Gensim \citep{vrehuuvrek2011gensim}. The latter are much less computationally expensive to obtain; 768-dimensional vectors were trained in one epoch in 1.61 days on a 16-CPU machine. CBOW embeddings are trained by using surrounding context to predict a center word. While training via CBOW has often shown inferior performance to training via skipgram (SG), this paper shows that with a proper implementation, performance of CBOW embeddings can be on par with SG. Our work puts the CBOW prediction objective to good use, harnessing it to predict anchored text in source language segments.

\section{Methodology}\label{se:methodology}

Neural MT systems have been shown by previous work \citep{bulte-tezcan-2019-neural} to be the state-of-the-art for FMR. In this article, we experiment on the one hand with word-based language models that are trained using context around a word, like those that use the continuous bag of words (CBOW) model \citep{mikolov2013efficient} (Word2Vec) or masked language modeling \citep{sanh2019distilbert} (BERT). On the other hand, it is our belief also that generative language modeling techniques may be a good candidate for accomplishing this task. To explore this avenue, we also compare output from these models with predictions obtained from prompting GPT-4 and find it to be competitive with the other methods. An example of a source sentence and the output from each method is provided in Table \ref{se:methodology:tbl:outputs} with predicted (or reference) word in bold. In our experiment, the two language modeling techniques as well as the generative approach are compared against machine translation and measured using character rate and accuracy against sets of anchored words from the test set. A prediction or translation was deemed correct when the center word from a tri-gram of anchored words was correctly found. In the following sub-sections, we discuss each approach. In Section \ref{se:experimental} we provide further details about the corpora and configuration.

\begin{table*}[!htbp]
\begin{center}
\begin{tabular}{|c|c|}
    \hline
    \textbf{Method} & \textbf{Output} \\
    \hline
     Original French & ``Afin d’\underline{évaluer si} l’établissement identifie toutes\\
     
     & les situations qui doivent être considérées comme\\
     
     & des défauts, conformément à l’article 178,\\
     
     & paragraphes 1 à 5, du règlement (UE) no 575/2013...''\\
     
     \hline
     Reference Translation & ``In order to \underline{assess whether the} institution\\
     
     & identifies all situations which are to be considered\\
     
     & defaults in accordance with Article 178 (1)\\
     
     & to (5) of Regulation (EU) No 575/2013...'' \\
     \hline
     Reference tri-gram & assess \textbf{whether} the \\
     \hline 
     BERT & assess \textbf{whether} the \\
     \hline
     Word2Vec & assess \textbf{commission} the \\
     \hline
     MT & assess \textbf{obligatory} the \\
     \hline
     GPT-4 & assess \textbf{and} the \\
     \hline
\end{tabular}
\caption{Anchored tri-gram reference and predictions (predicted word in bold)}
\label{se:methodology:tbl:outputs}
\end{center}
\end{table*}

\subsection{Machine Translation}\label{se:methodology:mt}

We train the neural MT system with Open-NMT \citep{klein2020opennmt} using the default transformer configuration. In order to get a wider range of difference with the MT system, we translate using two methods: (1) the translation of the $s'_{en}$ segment to $t^*_{fr}$ then translation from $t^*_{fr}$ to $s^*_{en}$; and, (2) the translation of the three-word sub-segment only (i.e. the anchored tri-gram with the center word to be translated) from $s_{trigram\textrm{-}en}'$ to $t^*_{trigram\textrm{-}fr}$ then translation from $t^*_{trigram\textrm{-}fr}$ to $s^*_{trigram\textrm{-}en}$.  For both methods, correctly translated center words from tri-grams were counted in the overall evaluation. Predictions by the other two methods were scored similarly. Further details on parameters and configuration are found in Section \ref{se:experimental:mt}.




\subsection{Word2Vec}\label{se:methodology:w2v}
\begin{figure}[H]
   \centering
    \includegraphics[width=0.38\textwidth]{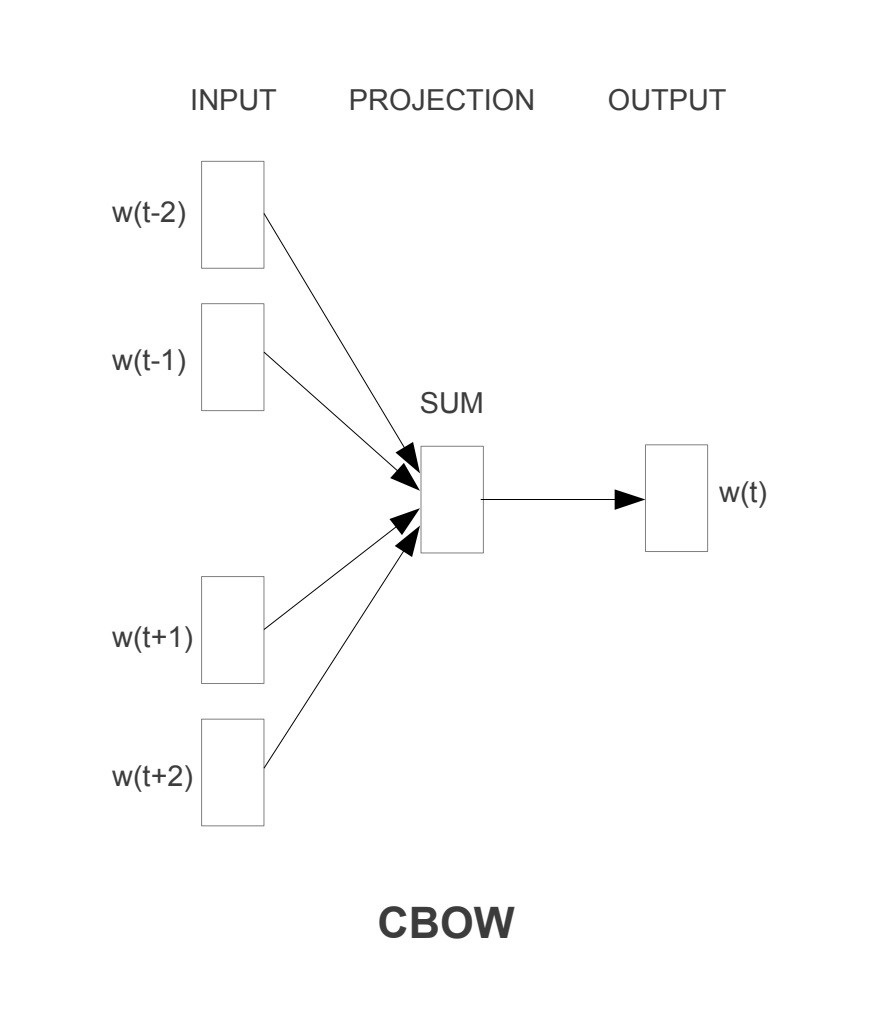}
    

   \caption{An illustration of predicting a word given the context around it (denoted as \textit{anchored words in this article)}, called Continuos Bag of Words (CBOW) by \protect \cite{mikolov2013efficient}.}
    \label{fig:cbow}
\end{figure}

We used a pre-trained language model (PLM) for experimentation with Word2Vec \citep{mikolov2013efficient}.\footnote{We use the pre-trained word news vectors from Google found here:\url{https://github.com/mmihaltz/word2vec-GoogleNews-vectors?tab=readme-ov-file.}} The hope is that through the use of a PLM we can capture context in several different domains, specifically the corpus that we use which is parliamentary in nature. 

The PLM weights from Word2Vec were used as a manner to predict anchored words due to the fact that the training method for them is based on a CBOW model.  CBOW was selected because, as shown in Figure \ref{fig:cbow}, its training objective most closely resembles the task we are trying to accomplish---the prediction of a word surrounded by anchored text (one word on the left and one word on the right). 

As a first step, the PLM was downloaded and experimented as-is in its out-of-the-box state which consists of 300 dimensions and a default vocabulary. Then, in order to fine-tune the model, we adapted it to our parliamentary corpus. After the fine-tuning of the model, anchored tri-grams were extracted from $s'$ and used as input to the PLM where the center word is used for prediction and the left and right ``anchors'' are used as input one-hot encoded embeddings, similar to the training exercise from \cite{mikolov2013efficient}. Further details on parameters and configuration are found in Section \ref{se:experimental:w2v}.

\subsection{BERT}\label{se:methodology:bert}

Models based on the BERT \citep{kenton2019bert} algorithm are used frequently in modern times. They use an attention mechanism \citep{vaswani2017attention} and are known to be capable of capturing information better than previous implementations such as Word2vec. Therefore, in order to compare both algorithms to MT for predicting anchored words, we experiment with DistilBERT \citep{sanh2019distilbert}, a BERT-based model that uses masked language modeling that in theory captures more parameters than the Word2Vec CBOW model.

Similar to the Word2Vec method, we fine-tune our DistilBERT model on the parliamentary corpus with a masked language modeling objective. We chose the masked language modeling objective as it is the most similar objective to CBOW. Further details on parameters and configuration are found in Section \ref{se:experimental:bert}.





\subsection{GPT-4}

We experiment with prompting GPT-4 to predict anchored text using a temperature of 0 and the following prompt: \textit{``You are an expert lexicographer and natural language processing assistant. Additionally, you are highly specialized in parliamentary proceedings. Given a trigram I provide with a '?' character in the center word, I need you to predict the '?' character with the most likely single-word token. Please return one predicted token without any text except the predicted token in your response. Do not provide the surrounding text or any additional information. Do not include the text 'predicting', 'predict', 'prediction', 'predicted' 'the predicted token is' or 'The predicted token is' in your response. Do not include any extra characters such as apostrophes, commas, colons, or semicolons in your response. Do not include any newline characters in your response.''}.

\section{Experimental Settings}\label{se:experimental}

\subsection{Corpus}\label{se:experimental:mt}

The corpus consists of 393,371 SL-TL pairs of European parliamentary proceedings, a freely available translation memory \citep{steinberger-etal-2012-dgt} obtained from the European Commission DGT-Translation Memory repository.\footnote{\url{https://joint-research-centre.ec.europa.eu/language-technology-resources/dgt-translation-memory_en}} The corpus is divided randomly with a random state of $42$. We divide the corpus up into 70\% train, 20\% dev, and 10\% test sets as shown in in Table \ref{se:experimental:tbl:splits}.

\begin{table*}[!htbp]
\begin{center}
\begin{tabular}{|lr|}
     \hline
      \textbf{Data set} & \textbf{Segment Size} \\
     \hline
     \textbf{Train} & 275,317 \\

     \textbf{Development} & 77,877 \\
     
     \textbf{Test} & 40,117 \\

     \hline
     
     \textbf{Total} & 393,371 \\
     
     \hline
\end{tabular}
\caption{Experimental sets from the European Commission DGT Translation memory used for creating and evaluating the three approaches.}
\label{se:experimental:tbl:splits}
\end{center}
\end{table*}

\subsection{Machine Translation}\label{se:experimental:mt}

As mentioned previously, we use the Open-NMT \citep{klein2020opennmt} framework to build our French to English (FR--EN) and English to French (EN--FR) MT system. The system is based on a transformer architecture model with the following hyper-parameters: A maximum sequence length of 500, an early stopping parameter of 4, 7,800 train steps, 1,000 validation steps, a bucket size of 262,144, a batch size of 4,096, and a validation batch size of 2,048. The optimizer is an Adam (beta2 of 0.998) optimizer with with fp16 activated, a learning rate of 2, noam decay, label smoothing of 0.1, a hidden size of 512, word vector size of 512, 8 attention heads, a dropout of 0.1, and an attention dropout of 0.1. The choice of parameter selection is inspired by previous work from Yasmin Moslem.\footnote{\url{https://github.com/ymoslem/OpenNMT-Tutorial}}

In order to verify that the NMT system is on-par with the latest MT systems for FR--EN and EN--FR, we first test the system in both directions on the test set. During test, we achieved a BLEU score of 55.84 for FR--EN and 62.60 for EN-FR. Nonetheless, as we show in Section \ref{se:results}, the translation of anchored words as measured by character rate and accuracy was not remarkable.


\subsection{Word2Vec}\label{se:experimental:w2v}

The CBOW algorithm for Word2vec is a well-known algorithm performed as a way of capturing semantics via a language model \citep{mikolov2013efficient}. We describe our Word2Vec CBOW implemenation. Before fine-tuning, the Word2Vec model has 300 dimensions with a window size of 2 and a minimum word count of 1. Additionally, pre-defined vocabulary is used in the Google News Vectors that contains billions of words. The model is fine-tuned with our training set which is tokenized using the NLTK RegexpTokenizer\footnote{\url{https://www.nltk.org/_modules/nltk/tokenize/regexp.html}}. The embeddings created from the training set use lockf at 1.0 and a window size of 3, similar to Zarrar Shehzad.\footnote{\url{https://czarrar.github.io/Gensim-Word2Vec/}}

\subsection{BERT}\label{se:experimental:bert}
Our BERT model is based on a PLM called DistilBERT\footnote{\url{https://github.com/huggingface/transformers/tree/main/examples/research_projects/distillation}}. \citep{sanh2019distilbert} We train DistilBERT using the HuggingFace PyTorch Trainer with 10 training epochs, a learning rate of 2e-5, weight decay of 0.01, and FP16 mixed precision set to true. Hyperparameters are inspired by HuggingFace.\footnote{\url{https://huggingface.co/learn/nlp-course/en/chapter7/3}}

\subsection{GPT-4}

GPT-4 was prompted using the gpt-4-turbo variant and queried repetitively through the OpenAI API. Due to newline mismatches that occurred during batch processing, we opted to run an API call for every line in the dataset.

\section{Results}\label{se:results}
\begin{table*}[!htb]
  \centering
  \begin{tabular}{lcccc}
    \hline
      & \textbf{60--69\%} &  \textbf{70--79\%} & \textbf{80--89\%} & \textbf{90--100\%} \\
    \hline
    BERT & \textbf{8.97} & \textbf{9.61} & \textbf{7.98} & \textbf{7.87} \\
    GPT  & 4.82 & 5.58 & 3.85 & 2.74 \\
    W2V & 3.39 & 3.46 & 2.89 & 3.02 \\
    NMT-1 & 0.15 & 0.14 & 0.28 & 0.19 \\
    NMT-2 & 3.75 & 4.36 & 4.16 & 6.35 \\
    \hline
  \end{tabular}
  \caption{Accuracy scores for various fuzzy-match threshold on five deep-learning approaches.}
   \label{sec:results:fmr_accuracy_dl}
\end{table*}

In this section, we compare the results obtained from running four approaches for predicting the anchored word: (1) Neural Machine Translation (NMT) (2) Word2Vec (3) BERT and (4) GPT-4. NMT is divided into the two approaches mentioned in Section \ref{se:methodology:mt} (sentence-level and tri-grams). Accuracy measurements are performed and reported for all holes\footnote{A hole is a span of a tri-gram where the center word is predicted.}. Additionally, we report on character matches for each approach after dividing the segments into fuzzy-match thresholds, common practice for FMR work (see \citep{ortega16a,bulte-tezcan-2019-neural}).

\begin{figure}[H]
   \centering
    \includegraphics[width=\linewidth]{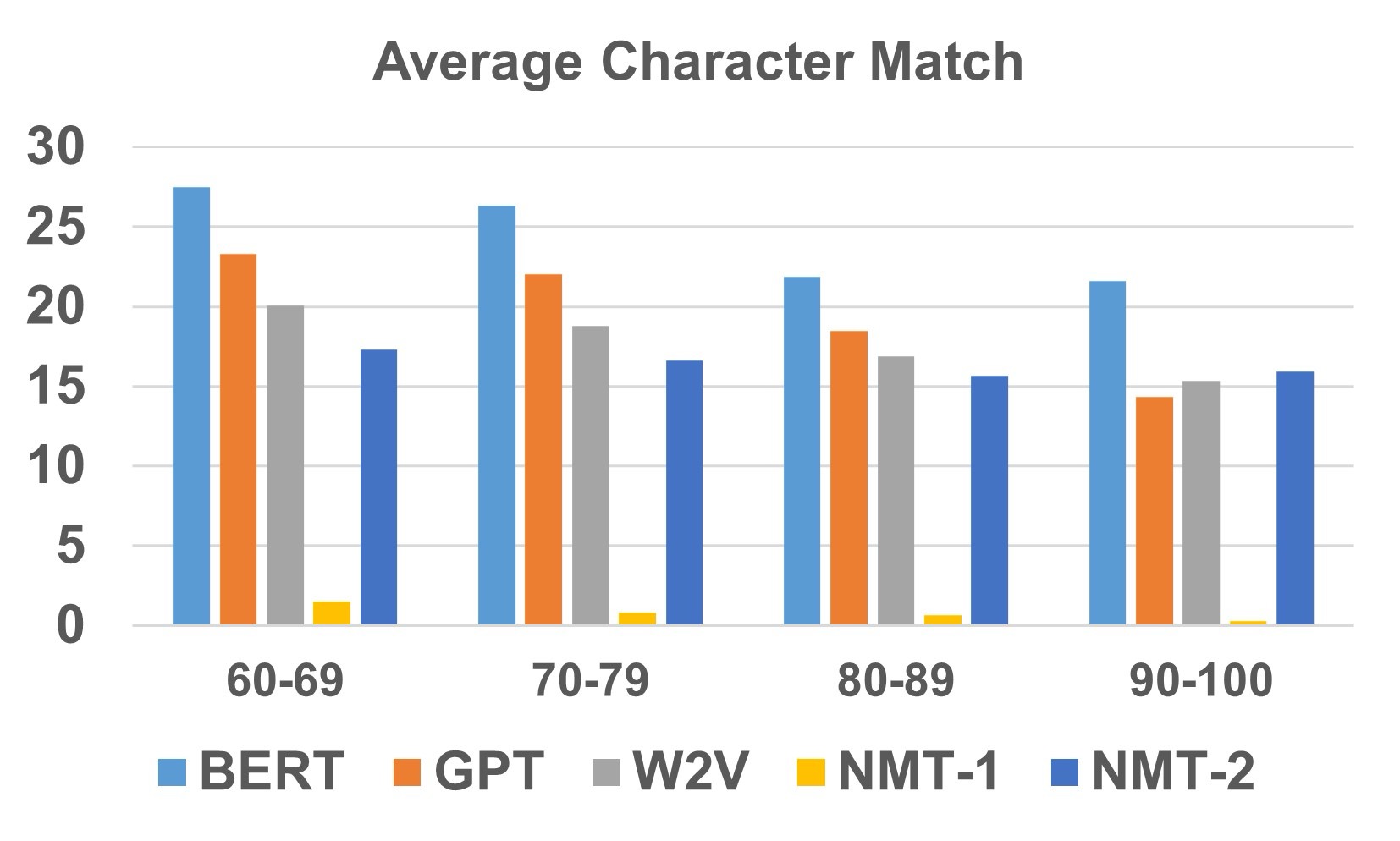}
    

   \caption{Average character match (y-axis) by fuzzy-match rate percentage (x-axis) by segment of the four experimental approaches: BERT, GPT, Word2Vec, Neural Machine Translation 1 and Neural Machine Translation 2 systems for different segment-level fuzzy-match thresholds.}
    \label{fig:character_comparison}
\end{figure}

First, we report on character match rates for the three approaches. Character match is defined as the number of characters in the output token that correspond to characters in the desired string. In Figure \ref{fig:character_comparison}, we report the average character match for GPT-4, BERT, Word2Vec, NMT-1 (segment-level MT) and NMT-2 (tri-gram MT). We observe a marked improvement in average character match with language modeling approaches (BERT and Word2Vec) and GPT-4 performs competitively in most cases. BERT outperforms all approaches across all fuzzy-match thresholds. From an MT standpoint, the secondary approach (called NMT-2 in Figure \ref{fig:character_comparison}) outperforms the primary approach; it appears that in our experiments the translation of anchored tri-grams is better than translating the entire segment.

Additionally, we measured the accuracy for the three approaches in order to better understand the hole span coverage. For accuracy, we measure only if prediction was correct or not; we do not take into account other predictions like blank, extra words, or others. To this end, we present accuracy scores in Table \ref{sec:results:fmr_accuracy_dl}.


In our experiments, we notice that the NMT systems perform better on stop words and digits such as the phrase: ``beyond \textbf{90} ghz''. Both the BERT and NMT systems were found to perform well in those situations. However, the MT system oftentimes did not replace one word only---in several cases it aggregated several words more. BERT performed well on average when compared with the other approaches. GPT remains competitive on all fuzzy match ranges except 90--100.

\section{Conclusion} \label{se:conclusion}

In this article, we have illustrated that via the use of a language model, predicting anchored words performed better in our experiments. The BERT model outperforms other approaches including neural machine translation (with two approaches) when measured via character match and tri-gram anchored word coverage. 

We also demonstrate how generative models might be prompted to aid in predicting anchored text. It is our belief that this work could assist CAT tools backed by TMs and MT systems.

\newpage

\begin{small}
\bibliographystyle{apalike}
\bibliography{amta2024}

\begin{thebibliography}{}

\bibitem[Achiam et~al., 2023]{achiam2023gpt}
Achiam, J., Adler, S., Agarwal, S., Ahmad, L., Akkaya, I., Aleman, F.~L., Almeida, D., Altenschmidt, J., Altman, S., Anadkat, S., et~al. (2023).
\newblock Gpt-4 technical report.
\newblock {\em arXiv preprint arXiv:2303.08774}.

\bibitem[Biccici and Dymetman, 2008]{biccici2008dynamic}
Biccici, E. and Dymetman, M. (2008).
\newblock Dynamic translation memory: Using statistical machine translation to improve translation memory fuzzy matches.
\newblock {\em Computational Linguistics and Intelligent Text Processing}, pages 454--465.

\bibitem[Bowker, 2002]{bowker2002computer}
Bowker, L. (2002).
\newblock {\em Computer-aided translation technology: a practical introduction}.
\newblock University of Ottawa Press.

\bibitem[Brown et~al., 1993]{brown93:tmo}
Brown, P.~F., Della~Pietra, S.~A., Della~Pietra, V.~J., and Mercer, R.~L. (1993).
\newblock The mathematics of statistical machine translation: Parameter estimation.
\newblock {\em Computational Linguistics}, 19(2):263--311.

\bibitem[Bulte and Tezcan, 2019]{bulte-tezcan-2019-neural}
Bulte, B. and Tezcan, A. (2019).
\newblock Neural fuzzy repair: Integrating fuzzy matches into neural machine translation.
\newblock In {\em Proceedings of the 57th Annual Meeting of the Association for Computational Linguistics}, pages 1800--1809, Florence, Italy.

\bibitem[Bulte et~al., 2018]{bulte18:_m3tra}
Bulte, B., Vanallemeersch, T., and Vandeghinste, V. (2018).
\newblock {M3TRA: integrating TM and MT for professional translators}.
\newblock In {\em Proceedings of the 21st Annual Conference of the European Association for Machine Translation}, pages 69--78, Alacant, Spain.

\bibitem[Dandapat et~al., 2011]{dandapat2011}
Dandapat, S., Morrissey, S., Way, A., and Forcada, M.~L. (2011).
\newblock Using example-based {MT} to support statistical {MT} when translating homogeneous data in a resource-poor setting.
\newblock In {\em Proceedings of the 15th Annual Conference of the European Association for Machine Translation}, pages 201--208, Leuven, Belgium.

\bibitem[Espla-Gomis et~al., 2011]{EsplGomis2011UsingWA}
Espla-Gomis, M., Sanchez-Martinez, F., and Forcada, M.~L. (2011).
\newblock Using word alignments to assist computer-aided translation users by marking which target-side words to change or keep unedited.
\newblock In {\em European Association for Machine Translation Conferences/Workshops}.

\bibitem[Gu et~al., 2018]{gu18}
Gu, J., Wang, Y., Cho, K., and Li, V.~O. (2018).
\newblock Search engine guided neural machine translation.
\newblock In {\em Proceedings of the 32 AAAI Conference on Artificial Intelligence}, pages 5133--5140, New Orleans, USA.

\bibitem[Hewavitharana et~al., 2005]{hewavitharana2005augmenting}
Hewavitharana, S., Vogel, S., and Waibel, A. (2005).
\newblock Augmenting a statistical translation system with a translation memory.
\newblock In {\em Proceedings of the 10th Annual Conference of the European Association for Machine Translation}, pages 126--132, Budapest, Hungary.

\bibitem[Irsoy et~al., 2020]{irsoy2020corrected}
Irsoy, O., Benton, A., and Stratos, K. (2020).
\newblock Corrected cbow performs as well as skip-gram.
\newblock {\em arXiv preprint arXiv:2012.15332}.

\bibitem[Kenton and Toutanova, 2019]{kenton2019bert}
Kenton, J. D. M.-W.~C. and Toutanova, L.~K. (2019).
\newblock Bert: Pre-training of deep bidirectional transformers for language understanding.
\newblock In {\em Proceedings of NAACL-HLT}, pages 4171--4186.

\bibitem[Klein et~al., 2020]{klein2020opennmt}
Klein, G., Hernandez, F., Nguyen, V., and Senellart, J. (2020).
\newblock The opennmt neural machine translation toolkit: 2020 edition.
\newblock In {\em Proceedings of the 14th Conference of the Association for Machine Translation in the Americas (Volume 1: Research Track)}, pages 102--109.

\bibitem[Koehn and Senellart, 2010]{koehn2010convergence}
Koehn, P. and Senellart, J. (2010).
\newblock Convergence of translation memory and statistical machine translation.
\newblock In {\em Proceedings of AMTA Workshop on MT Research and the Translation Industry}, pages 21--31, Denver, USA.

\bibitem[Kranias and Samiotou, 2004]{kranias04:_automatic}
Kranias, L. and Samiotou, A. (2004).
\newblock Automatic translation memory fuzzy match post-editing: a step beyond traditional {TM/MT} integration.
\newblock In {\em Proceedings of the {F}ourth {I}nternational {C}onference on {L}anguage {R}esources and {E}valuation}, pages 331--334, {L}isbon, {P}ortugal.

\bibitem[Levenshtein, 1966]{levenshtein1965binary}
Levenshtein, V. (1966).
\newblock Binary codes capable of correcting deletions, insertions and reversals.
\newblock {\em {S}oviet {P}hysics {D}oklady.}, 10(8):707--710.

\bibitem[Li et~al., 2016]{li16j}
Li, L., Parra~Escartin, C., and Liu, Q. (2016).
\newblock Combining translation memories and syntax-based {SMT}.
\newblock {\em Baltic Journal of Modern Computing}, 4:165--177.

\bibitem[Liu et~al., 2019]{liu19}
Liu, Y., Wang, K., Zong, C., and Su, K.-Y. (2019).
\newblock A unified framework and models for integrating translation memory into phrase-based statistical machine translation.
\newblock {\em Computer Speech and Language}, 54:176--206.

\bibitem[Mikolov et~al., 2013]{mikolov2013efficient}
Mikolov, T., Chen, K., Corrado, G., and Dean, J. (2013).
\newblock Efficient estimation of word representations in vector space.
\newblock {\em arXiv preprint arXiv:1301.3781}.

\bibitem[Ortega et~al., 2014]{ortega14a}
Ortega, J.~E., Sanchez-Martinez, F., and Forcada, M.~L. (2014).
\newblock Using any machine translation source for fuzzy-match repair in a computer-aided translation setting.
\newblock In {\em Proceedings of the 11th Biennial Conference of the Association for Machine Translation in the Americas (AMTA 2014, vol. 1: MT Rsearchers)}, pages 42--53, Vancouver, BC, Canada.

\bibitem[Ortega et~al., 2016]{ortega16a}
Ortega, J.~E., Sanchez-Martinez, F., and Forcada, M.~L. (2016).
\newblock Fuzzy-match repair using black-box machine translation systems: what can be expected?
\newblock In {\em Proceedings of the 12th Biennial Conference of the Association for Machine Translation in the Americas (AMTA 2016, vol. 1: MT Researchers' Track)}, pages 27--39, Austin, USA.

\bibitem[Sanh et~al., 2019]{sanh2019distilbert}
Sanh, V., Debut, L., Chaumond, J., and Wolf, T. (2019).
\newblock Distilbert, a distilled version of bert: smaller, faster, cheaper and lighter.
\newblock {\em arXiv preprint arXiv:1910.01108}.

\bibitem[Simard and Isabelle, 2009]{simard2009phrase}
Simard, M. and Isabelle, P. (2009).
\newblock Phrase-based machine translation in a computer-assisted translation environment.
\newblock In {\em Proceeding of the 12th Machine Translation Summit (MT Summit XII)}, pages 120--127, Quebec, Canada.

\bibitem[Steinberger et~al., 2012]{steinberger-etal-2012-dgt}
Steinberger, R., Eisele, A., Klocek, S., Pilos, S., and Schluter, P. (2012).
\newblock {DGT}-{TM}: A freely available translation memory in 22 languages.
\newblock In Calzolari, N., Choukri, K., Declerck, T., Dougan, M.~U., Maegaard, B., Mariani, J., Moreno, A., Odijk, J., and Piperidis, S., editors, {\em Proceedings of the Eighth International Conference on Language Resources and Evaluation (LREC'12)}, pages 454--459, Istanbul, Turkey. European Language Resources Association (ELRA).

\bibitem[Tezcan and Bulte, 2022]{Tezcan2022EvaluatingTI}
Tezcan, A. and Bulte, B. (2022).
\newblock Evaluating the impact of integrating similar translations into neural machine translation.
\newblock {\em Inf.}, 13:19.

\bibitem[Tezcan et~al., 2021]{tezcan2021towards}
Tezcan, A., Bulte, B., and Vanroy, B. (2021).
\newblock Towards a better integration of fuzzy matches in neural machine translation through data augmentation.
\newblock In {\em Informatics}, volume~8, page~7. MDPI.

\bibitem[Vaswani et~al., 2017]{vaswani2017attention}
Vaswani, A., Shazeer, N., Parmar, N., Uszkoreit, J., Jones, L., Gomez, A.~N., Kaiser, L., and Polosukhin, I. (2017).
\newblock Attention is all you need.
\newblock {\em Advances in neural information processing systems}, 30.

\bibitem[Vogel et~al., 1996]{vogel96:hbw}
Vogel, S., Ney, H., and Tillmann, C. (1996).
\newblock {HMM}-based word alignment in statistical translation.
\newblock In {\em COLING96}, pages 836--841, Copenhagen.

\bibitem[VRehuuVrek et~al., 2011]{vrehuuvrek2011gensim}
VRehuuVrek, R., Sojka, P., et~al. (2011).
\newblock Gensim—statistical semantics in python.
\newblock {\em Retrieved from genism. org}.

\bibitem[Wagner and Fischer, 1974]{Wagner:1974:SCP:321796.321811}
Wagner, R.~A. and Fischer, M.~J. (1974).
\newblock The string-to-string correction problem.
\newblock {\em Journal of the Association for Computing Machinery}, 21(1):168--173.

\bibitem[Zhechev and Genabith, 2010]{zhechev2010seeding}
Zhechev, V. and Genabith, J.~V. (2010).
\newblock Seeding statistical machine translation with translation memory output through tree-based structural alignment.
\newblock In {\em Proceedings of SSST-4 - 4th Workshop on Syntax and Structure in Statistical Translation}, pages 43--49, Dublin, Ireland.

\end{thebibliography}
\end{small}

\end{multicols}
\end{document}